\documentclass[letterpaper, 10 pt, conference]{ieeeconf}  

\IEEEoverridecommandlockouts                              

\usepackage{graphics}
\usepackage{graphicx}
\usepackage{tikz}
\usepackage{amsmath}
\usepackage[export]{adjustbox}
\usepackage{color}
\let\labelindent\relax
\usepackage{enumitem}
\usepackage{amsfonts}
\usepackage{subfig}
\usepackage{tabularx,booktabs}
\newcolumntype{C}{>{\centering\arraybackslash}X} 
\usepackage{acro}
\usepackage[percent,demo,abs]{overpic}
\usepackage[font=small]{caption}
\usepackage{hyperref}

\DeclareAcronym{cnn}{short = CNN, long = Convolutional Neural Network}
\DeclareAcronym{gan}{short = GAN, long = Generative Adversarial Network}
\DeclareAcronym{cgan_long}{short = cGAN, long = Conditional Generative Adversarial Network}
\DeclareAcronym{cgan}{short = cGAN, long = conditional GAN}
\DeclareAcronym{ss}{short = SS, long = Semantic Segmentation}

\title{\LARGE \bf
Empty Cities: Image Inpainting for a  Dynamic-Object-Invariant Space}

\author{Berta Bescos$^{1}$, Jos\'e Neira$^{1}$, Roland Siegwart$^{2}$ and Cesar Cadena$^{2}$
\thanks{This work has been supported by NVIDIA Corporation through the donation of a Titan X GPU, by the Spanish Ministry of Economy and Competitiveness (projects DPI2015-68905-P, FPI grant BES-2016-077836), the Arag\'on regional government (Grupo DGA T45-17R), the EU H2020 research project under grant agreement No 688652 and the Swiss State Secretariat for Education, Research and Innovation (SERI) No 15.0284.}
\thanks{$^{1}${Berta Bescos and Jos\'e Neira are with the Instituto de Investigaci\'on en Ingenier\'ia de Arag\'on (I3A), Universidad de Zaragoza, Zaragoza 50018, Spain
        {\tt\small \{bbescos,jneira\}@unizar.es}}}%
\thanks{$^{2}$Roland Siegwart and Cesar Cadena are with the Autonomous Systems Lab (ASL), ETH Z\"urich, Zurich 8092, Switzerland
        {\tt\small \{cesarc,rsiegwart\}@ethz.ch}}%
}

\begin{document}

\maketitle

\begin{abstract}

In this paper we present an end-to-end deep learning framework to turn images that show dynamic content, such as vehicles or pedestrians, into realistic static frames. 
This objective encounters two main challenges: detecting all the dynamic objects, and inpainting the static occluded background with plausible imagery. 
The second problem is approached with a conditional generative adversarial model that, taking as input the original dynamic image and its dynamic/static binary mask, is capable of generating the final static image. 
The former challenge is addressed by the use of a convolutional network that learns a multi-class semantic segmentation of the image. 

These generated images can be used for applications such as augmented reality or vision-based robot localization purposes. 
To validate our approach, we show both qualitative and quantitative comparisons against other state-of-the-art inpainting methods by removing the dynamic objects and hallucinating the static structure behind them.
Furthermore, to demonstrate the potential of our results, we carry out pilot experiments that show the benefits of our proposal for visual place recognition\footnote{All our code has been made available on \href{https://github.com/bertabescos/EmptyCities}{{\tt \small https://github.com/bertabescos/EmptyCities}}.}. 
%

\end{abstract}

\section{Introduction} 
\label{sec:intro}

Dynamic objects degrade the performance of vision-based robotic pose-estimation or localization tasks. 
The standard approach to deal with dynamic objects consists on detecting them in the images, and further classifying them as not valid information for such purposes. 
However, we propose to instead modify these images so that the dynamic content is converted realistically into static. 
We consider that the combination of experience and context allows us to hallucinate, \textit{i.e.}, inpaint, a geometrically and semantically consistent appearance of the static structure behind dynamic objects. 

\begin{figure} [h]
\vspace{-5mm}
\centering
\subfloat{\includegraphics[width=0.24\linewidth]{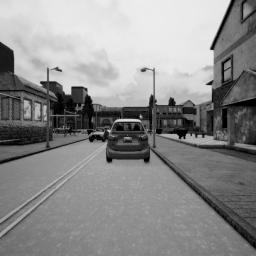}}
\hspace*{0.001\linewidth}
\subfloat{\includegraphics[width=0.24\linewidth]{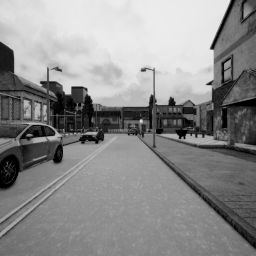}}
\hspace*{0.01\linewidth}
\subfloat{\includegraphics[width=0.24\linewidth]{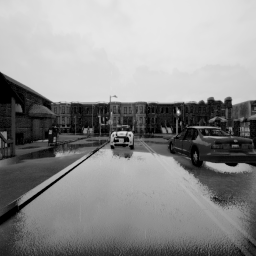}}
\hspace*{0.001\linewidth}
\subfloat{\includegraphics[width=0.24\linewidth]{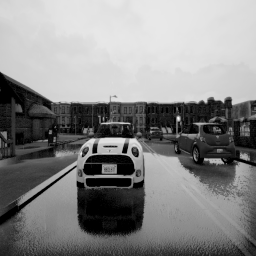}}
\\
\vspace*{-0.7\baselineskip}
\subfloat{\includegraphics[width=0.24\linewidth]{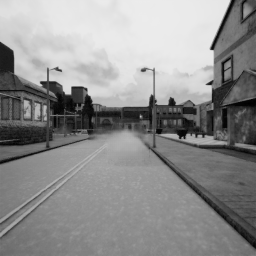}}
\hspace*{0.001\linewidth}
\subfloat{\includegraphics[width=0.24\linewidth]{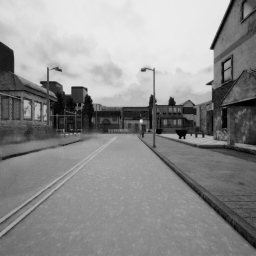}}
\hspace*{0.01\linewidth}
\subfloat{\includegraphics[width=0.24\linewidth]{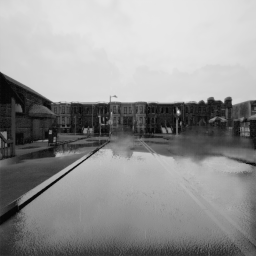}}
\hspace*{0.001\linewidth}
\subfloat{\includegraphics[width=0.24\linewidth]{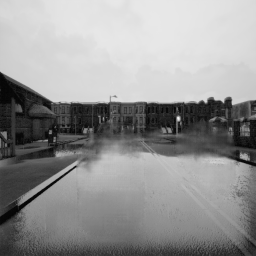}}
\caption{\label{fig:overview} Images at the same location with different dynamic content such as cars or pedestrians (top row) can be converted into the same static image, \textit{i.e.}, a dynamic-object-invariant space (bottom row).}
\vspace{-5mm}
\end{figure}
    
Turning images that contain dynamic objects into realistic static frames reveals several challenges:
\begin{enumerate}
\item{Detecting such dynamic content in the image. By this, we mean to detect not only those objects that are known to move such as vehicles, people and animals, but also the shadows and reflections that they might generate, since they also change the image appearance.}
\item{Inpainting the resulting space left by the detected dynamic content with plausible imagery. The resulting image would succeed in being realistic if the inpainted areas are both semantically and geometrically consistent with the static content of the image.}
\end{enumerate}

The former challenge can be addressed with geometrical approaches if a sequence of images is available. 
This procedure usually consists on studying the optical flow consistency along the images~\cite{wang2014motion, alcantarilla2012combining}. 
In the case in which only one frame is available, deep learning is the approach that excels at this task by the use of \acp{cnn}~\cite{he2017mask,romera2018erfnet}. 
These frameworks have to be trained with the previous knowledge of what classes are dynamic and which ones are not. 
Recent works show that it is possible to acquire this knowledge in a self-supervised manner~\cite{barnes2017driven, zhou2018dynamic}.

Regarding the second challenge, recent image inpainting approaches that do not use deep learning use image statistics  of the remaining  image to fill in the holes. The work of Telea~\cite{telea2004image} estimates the pixel value with the normalized weighted sum of all the known pixels in the neighbourhood. While this approach generally produces smooth results, it is limited by the available image statistics and has no concept of visual semantics. However, neural networks learn semantic priors and meaningful hidden representations in an end-to-end fashion, which have been used for recent image inpainting efforts~\cite{liu2018image,yu2018generative,iizuka2017globally,pathak2016context}. These networks employ convolutional filters on images, replacing the removed content with inpainted areas that usually have both geometrical and semantic consistency with the rest of the image.    

Both challenges can also be seen as one single task: translating a dynamic image into a corresponding static image. In this direction, Isola \textit{et~al.}~\cite{isola2017image} propose a general-purpose solution for image-to-image translation.
    
In this paper we present an end-to-end deep learning framework to turn images that have dynamic content into realistic static frames. 
This can be used for augmented reality, cinematography, and vision-based localization tasks. 
It could also be of interest for the creation of high-detail road maps, as well as for street-view imagery suppliers as a privacy measure to replace faces and license plates blurring.
    
Just like Isola \textit{et~al.}~\cite{isola2017image} succeed in translating images from day to night, aerial to map view, sketches to photos, \textit{etc.}, our paper builds on their work to translate images from a dynamic space into a static one. The main difference between our objective and his is that, while they apply the same translation to the whole image, we keep the static areas of the image almost untouched, and translate the dynamic parts into static ones. We have adapted their framework to our specific task by introducing a new loss that, combined with the integration of a semantic segmentation network achieves the final objective of creating a dynamic-object-invariant space. 
An example of our pipeline results can be seen in Fig.~\ref{fig:overview}. 

\section{Related Work}
\label{sec:related}
Previous works have attempted to reconstruct the background occluded by dynamic objects in the images with information from previous frames \cite{bescos2018dynaslam, scona2018staticfusion, granados2012background, uittenbogaard2018moving}.
If only one frame is available, the occluded background can only be reconstructed by image inpainting techniques.

\textbf{Image Inpainting.}
Among the non-learning approaches to image inpainting, propagating appearance information from neighboring pixels to the target region is the usual procedure~\cite{telea2004image}.
Accordingly, these methods succeed in dealing with narrow holes, where color and texture vary smoothly, but fail when handling big holes, resulting in over-smoothing. 
Differently, patch-based methods~\cite{efros2001image} operate by iteratively searching for relevant patches from the image non-hole regions. 
These approaches are computationally expensive and therefore not fast enough for real-time applications. 
Moreover, they do not make semantically aware patch selections.

Deep learning based methods usually initialize the image holes with a constant value, and further pass it through a \ac{cnn}. 
Context Encoders \cite{pathak2016context} were among the first ones to successfully use a standard pixel-wise reconstruction loss, as well as an adversarial loss for image inpainting tasks. 
Due to the resulting artifacts, Yang \textit{et~al.}~\cite{yang2017high} take the result from Context Encoders as input and then propagates the texture information from non-hole regions to fill the hole regions as post-processing.
Song \textit{et~al.}~\cite{song2017image} use a refinement network in which a blurry initial hole-filling result is used as the input, then iteratively replaced with patches from the closest non-hole regions in the feature space. 
Iizuka \textit{et~al.}~\cite{iizuka2017globally} extend Content Encoders by defining both global and local discriminators, then apply a post-processing. 
Following this work, Yu \textit{et~al.}~\cite{yu2018generative} replaced the post-processing with a refinement network powered by the contextual attention layers. 
The recent work of Liu \textit{et~al.}~ \cite{liu2018image} obtains amazing inpainting results by using partial convolutions. 

In contrast, the work by Ulyanov \textit{et~al.}~\cite{ulyanov2017deep} proves that there is no need for external dataset training. 
The generative network itself can rely on its structure to complete the corrupted image. 
However, this approach usually applies several iterations ($\sim$50000) to get good and detailed results.

\begin{figure} [t]
\vspace{-1mm}
\centering
\includegraphics[width=0.87\linewidth]{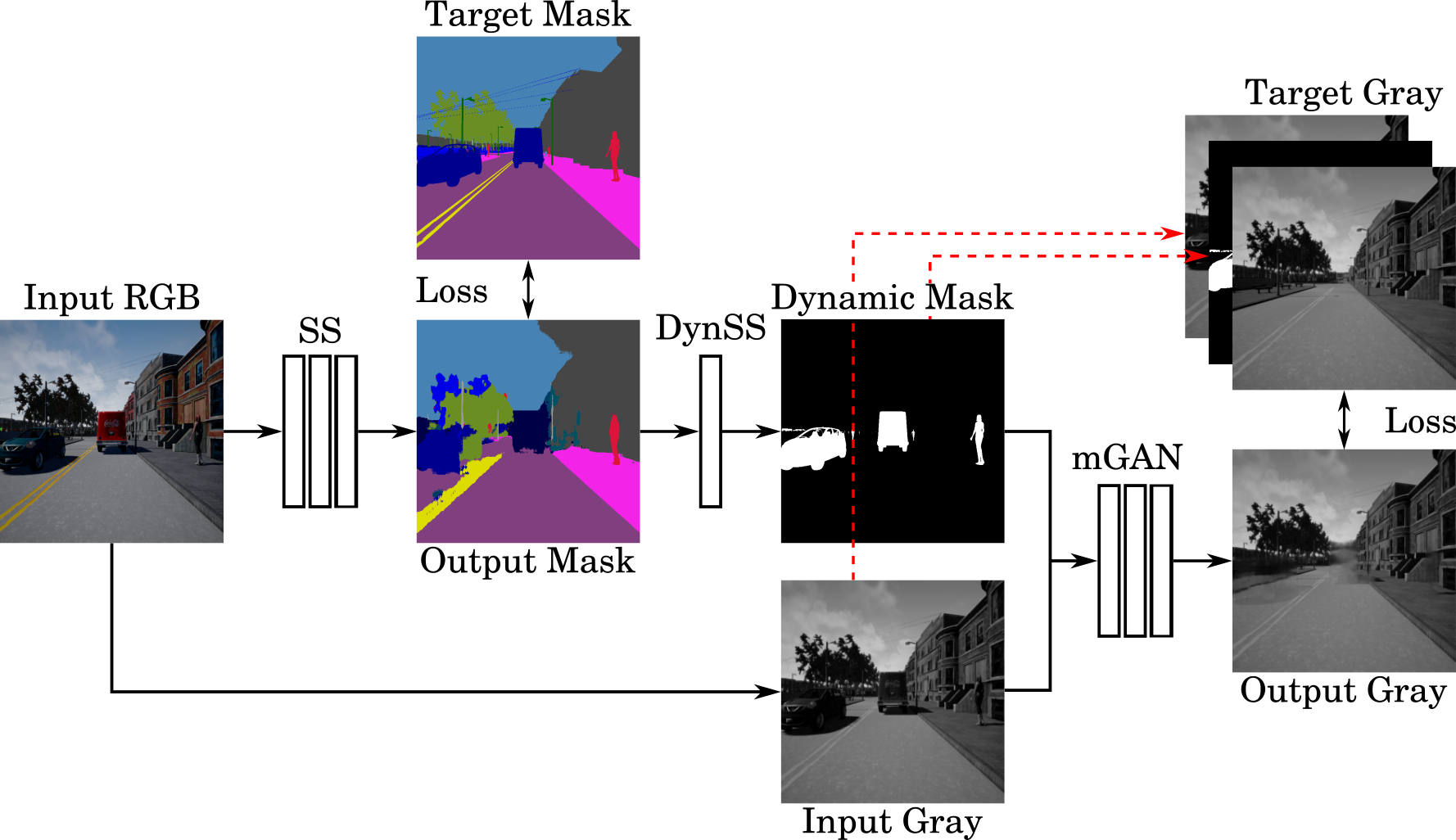}
\caption{\label{fig:Pipeline} Block diagram of our proposal. 
We first compute the segmentation of the RGB dynamic image, as well as its loss against the ground-truth. 
Then, the dynamic/static binary mask is obtained, and used together with the dynamic image to compute the static image. 
Its loss is back-propagated until the RGB dynamic image.}
\vspace{-5mm}
\end{figure}

Our work does not perform pure inpainting but image-to-image translation with the help of a mask, coming from a semantic segmentation network. 
This means that we cannot initialize the ``holes'' with any placeholder values since we do not want to learn that pixels with this particular value have to be transformed. 
In our case, our input consists of the dynamic original image with the dynamic/static mask concatenated. 
Different from the other approaches, we perform this task in gray scale instead of in RGB. 
The motivation for this is that learning a mapping from 1D$\rightarrow$1D instead of from 3D$\rightarrow$3D is simpler and therefore leads to having less room for wrong reconstructions. 
In addition, many visual localization applications only need the images grayscale information.
Still, as future work, we consider including a RGB version.
Moreover, note that using the image-to-image translation approach allows us to slightly modify the image non-hole regions for better accommodation of the reconstructed areas.


\section{System Description}
\label{sec:system}

Fig.~\ref{fig:Pipeline} shows an overview of our system during training time. 
First of all, we obtain the pixel-wise semantic segmentation of the RGB dynamic image (SS) and we compute its loss against the ground-truth. 
Then, the segmentation of only the dynamic objects is obtained with the convolutional layer DynSS. 
Once we have this dynamic/static binary mask, we convert the RGB dynamic image to gray scale and compute the static image, also in gray scale, with the use of a U-Net, which has been trained in an adversarial way (mGAN). 
The loss of this generated image, in addition to the appearance $L1$ loss, are back-propagated until the RGB dynamic image, together with the previous computed semantic segmentation loss. 
All the different stages, as well as the ground-truth generation, are described in subsections \ref{subsec:data} to \ref{subsec:dynSS}.

\subsection{Data Generation}
\label{subsec:data}
We have explored our method using CARLA~\cite{Dosovitskiy17}. 
CARLA is an open-source simulator for autonomous driving research, that provides open digital assets --urban layouts, buildings, vehicles, pedestrians, \textit{etc.}-- and supports flexible specification of sensor suites and environmental conditions. 
We have generated over 12000 image pairs consisting of a target image captured with neither vehicles nor pedestrians, and a corresponding input image captured at the same pose with the same illumination conditions, but with cars, tracks and people moving around. 
These images have been recorded using a front and a rear RGB camera mounted on a car. 
Their ground-truth semantic segmentation has also been captured. 
By manually selecting those dynamic classes (vehicles and pedestrians), we can easily obtain the ground-truth dynamic/static segmentation too.
CARLA offers two different towns that we have used for training and testing, respectively. 
Our dataset, together with more information about our framework, is available on \href{https://bertabescos.github.io/EmptyCities/}{{\tt \small https://bertabescos.github.io/EmptyCities/}}.

At present, we are limited to training on synthetic datasets since, to our knowledge, no real-world dataset exists that provides RGB images captured under same illumination conditions at identical poses over long trajectories, with and without dynamic objects. 
Recording a dataset ourselves would require huge amounts of both time and resources.

\subsection{Dynamic-to-Static Translation}

A \ac{gan} is a generative model that learns a mapping from a random noise vector $z$ to an output image~$y$, \mbox{$G$: $z \rightarrow y$}~\cite{goodfellow2014generative}. In contrast, a \ac{cgan} learns a mapping from observed image $x$ and optional random noise vector $z$, to $y$, \mbox{$G$ : $ \lbrace x,z \rbrace \rightarrow y $}~\cite{gauthier2014conditional}, or \mbox{$G$ : $ x \rightarrow y $}~\cite{isola2017image}. The generator $G$ is trained to produce outputs indistinguishable from ``real'' images by an adversarially trained discriminator $D$, which is trained to do as well as possible at detecting the generator's ``fakes''.

The objective of a \ac{cgan} can be expressed as 
\begin{multline}\label{eq:cGAN}
\mathcal{L}_{cGAN}(G,D)  = \mathbb{E}_{x,y}[\log{D(x,y)}] + \\
\mathbb{E}_{x}[\log{(1-D(x,G(x)))}],
\end{multline}

where $G$ tries to minimize this objective against an adversarial $D$ that tries to maximize it. Previous approaches have found it beneficial to mix the \ac{gan} objective with a more traditional loss, such as $L1$ or $L2$ distance \cite{pathak2016context}. The discriminator's job remains unchanged, but the generator is tasked not only with fooling the discriminator, but also with being near the ground-truth in a $L1$ sense, as expressed in
\begin{equation} \label{eq:L1}
G^*  = \text{arg}\,\min\limits_{G} \max\limits_{D}  \mathcal{L}_{cGAN}(G,D) + \lambda_{1} \cdot \mathcal{L}_{L1}(G),
\end{equation}

where $\mathcal{L}_{L1}(G)  = \mathbb{E}_{x,y}[||y-G(x)||_1]$.
The recent work of Isola \textit{et~al.}~\cite{isola2017image} shows that \ac{cgan}s are suitable for image-to-image translation tasks, where the output image is conditioned on its corresponding input image, \textit{i.e.}, it translates an image from one space into another (RGB appearance to drawings, day to night, \textit{etc.}). 
The main difference between our objective and his is that, while they apply the same mapping to the whole image, we want to keep almost untouched the static areas of the input image, and we want to translate the dynamic parts into plausible static ones. 
This problem could also be seen as inpainting. 
However, our method differs in that, in addition to changing the content of the image hole regions, it might also change the non-hole areas for a more realistic output (for example, dynamic objects shadows could also be removed even if unmasked). 

It is well known that $L2$ and $L1$ losses produce blurry results on image generation problems, \textit{i.e.}, they can capture the low frequencies but fail to encourage high frequency crispness. 
This motivates restricting the \ac{gan} discriminator to only model high frequency structures. 
Following this idea, Isola \textit{et~al.}~\cite{isola2017image} adopt a discriminator architecture that classifies each $N \times N$ patch in an image --rather than classifying the image as a whole-- as real or fake. 

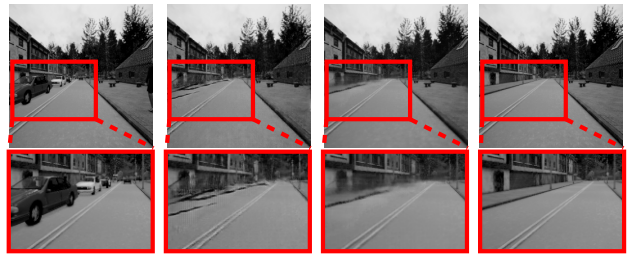
\begin{figure} [t]
\vspace{-3mm}
\centering
\subfloat{
\hspace*{-0.03\linewidth}
\begin{tikzpicture}
    \node[anchor=south west,inner sep=0] (image) at (0,0) {\includegraphics[width=0.22\linewidth]{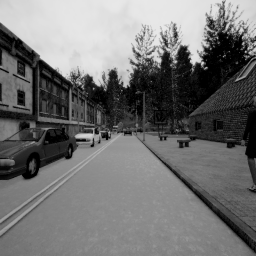}};
    \begin{scope}[x={(image.south east)},y={(image.north west)}]
        \draw[red,ultra thick] (0.02,0.20) rectangle (0.60,0.60);
        \foreach \i/\j in {{(0.02,0.20)/(0.0,0.0)},{(0.60,0.20)/(1.0,0.0)}}
            \draw [red,ultra thick,dashed] \i -- \j;
    \end{scope}
\end{tikzpicture}}
\hspace*{0.001\linewidth}
\subfloat{
\hspace*{-0.03\linewidth}
\begin{tikzpicture}
    \node[anchor=south west,inner sep=0] (image) at (0,0) {\includegraphics[width=0.22\linewidth]{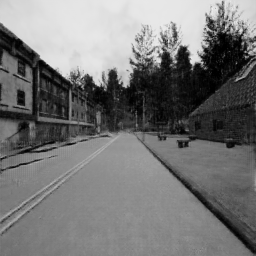}};
    \begin{scope}[x={(image.south east)},y={(image.north west)}]
        \draw[red,ultra thick] (0.02,0.20) rectangle (0.60,0.60);
        \foreach \i/\j in {{(0.02,0.20)/(0.0,0.0)},{(0.60,0.20)/(1.0,0.0)}}
            \draw [red,ultra thick,dashed] \i -- \j;
    \end{scope}
\end{tikzpicture}}
\hspace*{0.001\linewidth}
\subfloat{
\hspace*{-0.03\linewidth}
\begin{tikzpicture}
    \node[anchor=south west,inner sep=0] (image) at (0,0) {\includegraphics[width=0.22\linewidth]{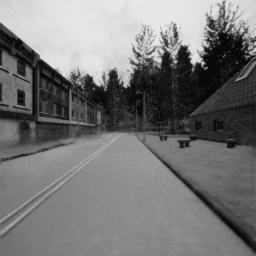}};
    \begin{scope}[x={(image.south east)},y={(image.north west)}]
        \draw[red,ultra thick] (0.02,0.20) rectangle (0.60,0.60);
        \foreach \i/\j in {{(0.02,0.20)/(0.0,0.0)},{(0.60,0.20)/(1.0,0.0)}}
            \draw [red,ultra thick,dashed] \i -- \j;
    \end{scope}
\end{tikzpicture}}
\hspace*{0.001\linewidth}
\subfloat{
\hspace*{-0.03\linewidth}
\begin{tikzpicture}
    \node[anchor=south west,inner sep=0] (image) at (0,0) {\includegraphics[width=0.22\linewidth]{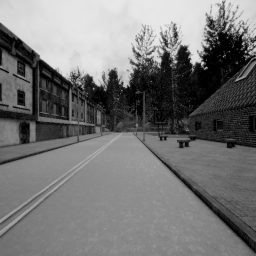}};
    \begin{scope}[x={(image.south east)},y={(image.north west)}]
        \draw[red,ultra thick] (0.02,0.20) rectangle (0.60,0.60);
        \foreach \i/\j in {{(0.02,0.20)/(0.01,0.01)},{(0.60,0.20)/(1.0,0.0)}}
            \draw [red,ultra thick,dashed] \i -- \j;
    \end{scope}
\end{tikzpicture}}
\\
\vspace*{-0.85\baselineskip}
\setcounter{subfigure}{0}
\subfloat[\label{fig:disc_in} Input]{
\hspace*{-0.03\linewidth}
\begin{tikzpicture}
    \node[anchor=south west,inner sep=0] (image) at (0,0) 
    {\adjincludegraphics[trim={{.02\width} {.20\height} {.40\width} {.40\height}}, clip, width=0.22\linewidth]{Images/input_947.png}};
    \begin{scope}[x={(image.south east)},y={(image.north west)}]
        \draw[red,ultra thick] (0.0,0.0) rectangle (1,1);
    \end{scope}
\end{tikzpicture}}
\hspace*{0.001\linewidth}
\subfloat[\label{fig:disc_nm} cGAN]{
\hspace*{-0.03\linewidth}
\begin{tikzpicture}
    \node[anchor=south west,inner sep=0] (image) at (0,0) 
    {\adjincludegraphics[trim={{.02\width} {.20\height} {.40\width} {.40\height}}, clip, width=0.22\linewidth]{Images/output_947_disc.png}};
    \begin{scope}[x={(image.south east)},y={(image.north west)}]
        \draw[red,ultra thick] (0.0,0.0) rectangle (1,1);
    \end{scope}
\end{tikzpicture}}
\hspace*{0.001\linewidth}
\subfloat[\label{fig:disc_m} mGAN]{
\hspace*{-0.03\linewidth}
\begin{tikzpicture}
    \node[anchor=south west,inner sep=0] (image) at (0,0) 
    {\adjincludegraphics[trim={{.02\width} {.20\height} {.40\width} {.40\height}}, clip, width=0.22\linewidth]{Images/output_947.png}};
    \begin{scope}[x={(image.south east)},y={(image.north west)}]
        \draw[red,ultra thick] (0.0,0.0) rectangle (1,1);
    \end{scope}
\end{tikzpicture}}
\hspace*{0.001\linewidth}
\subfloat[\label{fig:disc_gt} Ground-truth]{
\hspace*{-0.03\linewidth}
\begin{tikzpicture}
    \node[anchor=south west,inner sep=0] (image) at (0,0) 
    {\adjincludegraphics[trim={{.02\width} {.20\height} {.40\width} {.40\height}}, clip, width=0.22\linewidth]{Images/target_947.png}};
    \begin{scope}[x={(image.south east)},y={(image.north west)}]
        \draw[red,ultra thick] (0.0,0.0) rectangle (1,1);
    \end{scope}
\end{tikzpicture}}
\caption{\label{fig:disc} Qualitative results of the improvements achieved by conditioning the discriminator on both input and dynamic/static mask \protect\subref{fig:disc_m}, instead of only on the input \protect\subref{fig:disc_in}, as c\ac{gan}s do~\protect\subref{fig:disc_nm}.} 
\vspace{-5mm}
\end{figure}

\begin{figure*} [h]
\vspace{-5mm}
\centering
\subfloat{\includegraphics[height=.19\linewidth]{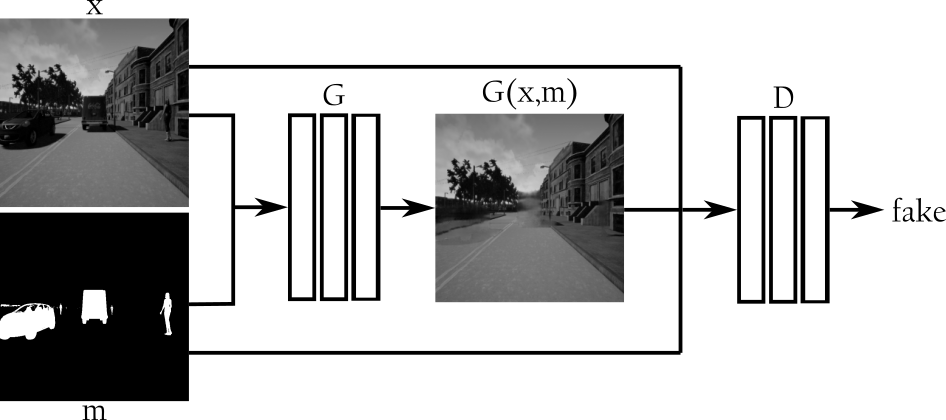}}
\hspace{0.5cm}
\centering
\subfloat{\includegraphics[height=.19\linewidth]{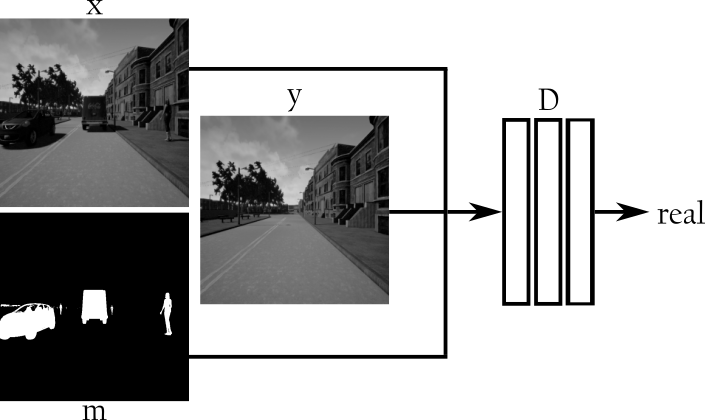}}
\caption{\label{fig:CMGAN} The discriminator $D$ has to learn to differ between the real images ($y$) and the images produced by the generator ($G(x,m)$). $D$ is conditioned by both the dynamic/static binary mask $m$ and the input image $x$ to make a better decision about the origin of the image.}
\vspace{-5mm}
\end{figure*}

For our objective, object masks are specially considered to re-formularize the training objectives. 
We adopt a variant of the \ac{cgan} that we call mGAN. 
mGANs learn a mapping from observed image $x$ and binary mask $m$, to $y$, \mbox{$G$ : $ \lbrace x,m \rbrace \rightarrow y $}. 
When applying this, we see that the dynamic objects in the image have been inpainted with high frequency texture but there are many artifacts (see Fig.~\ref{fig:disc_nm}). 
One of the reasons is that, in most of the training images the relationship between the static and dynamic regions sizes is unbalanced, \textit{i.e.}, static regions occupy usually a much bigger area. 
This leads us to believe that the influence of dynamic regions on the discriminator response is significantly reduced.
As a solution to this problem, we propose to change the discriminator loss so that there is more emphasis on the main areas that have to be inpainted, according to
\begin{multline} \label{eq:Dm}
\mathcal{L}_{mGAN}(G,D)  = \mathbb{E}_{x,y}[\log{D_{m}(x,m,y)}] + \\
\mathbb{E}_{x}[\log{(1-D_{m}(x,m,G(x,m)))}],
\end{multline}

where $D_{m}(x,m,y) = D(x,y) \otimes(1+m \cdot (\gamma-1))$.
The operator $\otimes$ means the element-wise matrix product, and the parameter $\gamma$ is a scalar that has been set to $2$.
A greater $\gamma$ value leads to better inpainting results in the masked areas, but the quality of the unmasked ones is compromised.
A smaller $\gamma$ value has very little effect on the results with regard to the original discriminator setup.
A good trade-off between the emphasis given to the masked compared to the unmasked regions is obtained with $\gamma = 2$. 
Fig.~\ref{fig:disc_nm} shows our output if the discriminator is conditioned only on the input, in contrast with the dicriminator conditioned on both the input and the mask (Fig.~\ref{fig:disc_m}).
The last one shows more realistic results. 
This training procedure is diagrammed in Fig.~\ref{fig:CMGAN}.

\subsection{Semantic Segmentation}
\ac{ss} is a challenging task that addresses many of the perception needs of intelligent vehicles in a unified way. 
Deep neural networks excel at this task, as they can be trained end-to-end to accurately classify multiple object categories in an image at pixel level. 
However, few architectures have a good trade-off between high quality and computational resources. 
The recent work of Romera \textit{et~al.}~\cite{romera2018erfnet} (ERFNet) uses residual connections to remain efficient while retaining remarkable accuracy.

Romera \textit{et~al.}~\cite{romera2018erfnet} have made public some of their trained models \cite{erfnet2017er}. 
We use for our approach the ERFNet model with encoder and decoder both trained from scratch on Cityscapes train set \cite{cordts2016cityscapes}. 
We have fine tuned their model to adjust it to our inpainting approach by back-propagating the loss of the semantic segmentation $\mathcal{L}_{CE}(SS)$, calculated using the class weights they suggest, $w$, and the adversarial loss of our final inpainted model $\mathcal{L}_{mGAN}(G,D)$. 
The \ac{ss} network's job can be therefore expressed as:
\begin{equation}
\label{eq:LossMask}
SS^{*} = \text{arg}\,\min\limits_{SS} \max\limits_{D}
\mathcal{L}_{mGAN}(G,D) + \lambda_2 \cdot \mathcal{L}_{CE}(SS),
\end{equation}

where $\mathcal{L}_{CE}(SS) = w[class] \cdot (log(\sum_{j} \exp(y_{SS}[j])) - y_{SS}[class])$. 
Its objective is to produce an accurate semantic segmentation $y_{SS}$, but also to fool the discriminator $D$.

\subsection{Dynamic Objects Semantic Segmentation}
\label{subsec:dynSS}
Once the semantic segmentation of the RGB image is done, we can select those classes known to be dynamic (vehicles and pedestrians). 
This has been done by applying a \textit{SoftMax} layer, followed by a convolutional layer with a kernel of $n \times 1 \times 1$, where $n$ is the number of classes, and with the weights of those dynamic and static channels set to $w_{dyn}$ and $w_{stat}$ respectively. 
$w_{dyn}$ and $w_{stat}$ are calculated following $w_{dyn} = \frac{n-n_{dyn}}{n}$ and $w_{stat} = -\frac{n_{dyn}}{n}$, where $n_{dyn}$ stands for the number of dynamic existing classes.

The consequent output passes through a $Tanh$ layer to obtain the wanted dynamic/static mask. 
Note that the defined weights $w_{dyn}$ and $w_{stat}$ are not changed during training time.

A possible extension of this work would contain a greater list of dynamic objects, such as construction sites, posters or temporary festival booths. 
This is not included at this time due to the lack of availability of training data.
	

\section{Experimental Results}
\label{sec:result}
\subsection{Main Contributions}
Here we report the improvements achieved by using, for our particular case, gray-scale instead of RGB images.
We also show how the error drops down when using a generator $G$ that learns a mapping from observed image $x$ and binary mask $m$ to $y$, $G$:~$ \lbrace x,m \rbrace \rightarrow y $, instead of a mapping from image $x$ to $y$, $G$:~$ x \rightarrow y $. 
Furthermore, we report how conditioning the discriminator on both the input and the binary mask $D(x,m,y)$, instead of on only the input $D(x,y)$, helps getting better results, see Table~\ref{tab:variants}.
%

\begin{table}[t]
\vspace{+2mm}
\begin{tabularx}{\linewidth}{@{}l*{5}{C}}
\toprule
Experiment & $G(x)_{\scalebox{1}{$\scriptscriptstyle RGB$}}$ & $G(x)$ & $G(x,m)$ & $G(x,m)$ \\
& $D(x,y)_{\scalebox{1}{$\scriptscriptstyle RGB$}}$ & $D(x,y)$ & $D(x,y)$ & $D(x,m,y)$ \\
\midrule
$L1 (\%)$ & 2.27 & 1.87 & 1.21 & \textbf{0.97} \\
$L1_{in} (\%)$ & 9.87 & 9.17 & 6.69 & \textbf{6.00} \\
$L1_{out} (\%)$ & 2.02 & 1.61 & 1.00 & \textbf{0.78} \\
\bottomrule
\end{tabularx}
\caption{\label{tab:variants} Quantitative evaluations of the performance of our contributions in the inpainting task on the test synthetic images.}
\vspace{-6mm}
\end{table}

The existence of many possible solutions renders difficult to define a metric to evaluate image inpainting~\cite{yu2018generative}.
Nevertheless, we follow previous works and report the $L1$ error.
Using RGB images usually leads to obtaining inpainting efforts on colorful areas such as cars, but also road signals and traffic lights, that we certainly want to keep untouched.
The main improvement carried out by working in gray scale is in the unmasked areas $L1_{out}$.
By using the mask to train both the generator $G$ and the discriminator $D$, we obtain a more accurate static-to-dynamic translation in both the masked ($L1_{in}$) and unmasked regions.
Results reported from now on are obtained with $G(x,m)$ and $D(x,m,y)$, \textit{i.e.}, mGANs.

\begin{figure*} [t]
\vspace{-5mm}
\centering
\subfloat{\begin{overpic}[width=.161\linewidth]{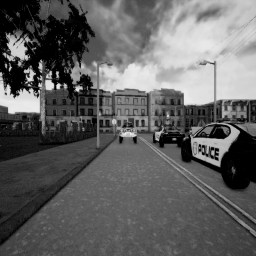}\put(1,56.5){\includegraphics[scale=0.04]{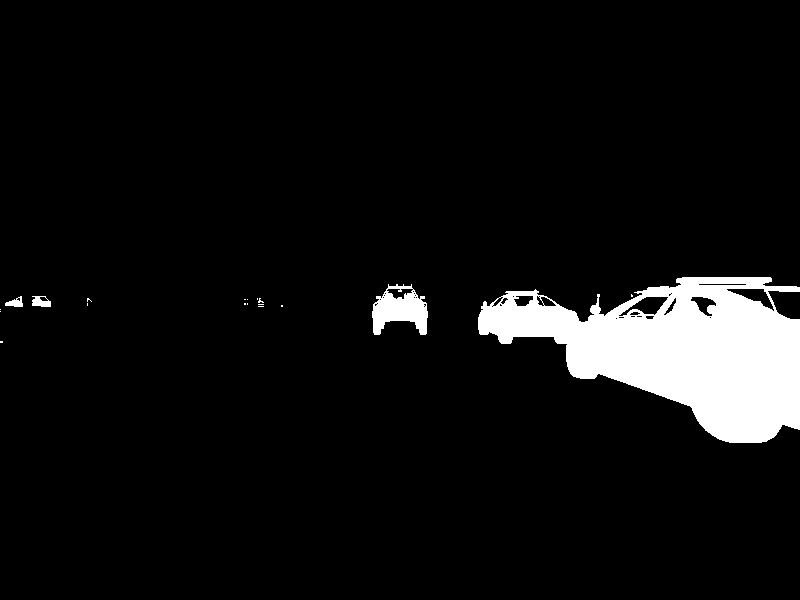}}\end{overpic}}
\hspace{\fill}
\subfloat{\includegraphics[width=.161\linewidth]{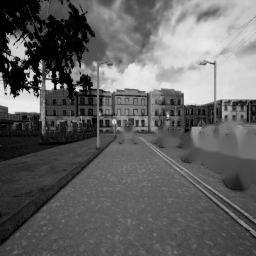}}
\hspace{\fill}
\subfloat{\includegraphics[width=.161\linewidth]{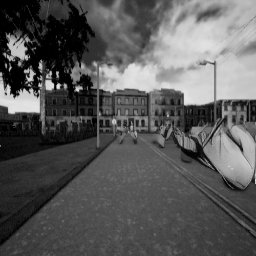}}
\hspace{\fill}
\subfloat{\includegraphics[width=.161\linewidth]{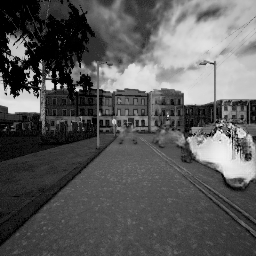}}
\hspace{\fill}
\subfloat{\includegraphics[width=.161\linewidth]{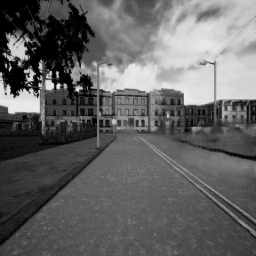}}
\hspace{\fill}
\subfloat{\includegraphics[width=.161\linewidth]{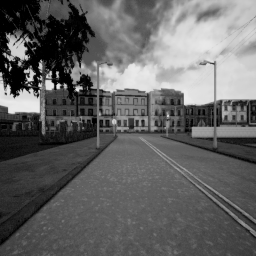}}
\\
\vspace*{-0.6\baselineskip}
\subfloat{\begin{overpic}[width=.161\linewidth]{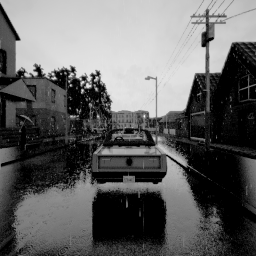}\put(1,56.0){\includegraphics[scale=0.04]{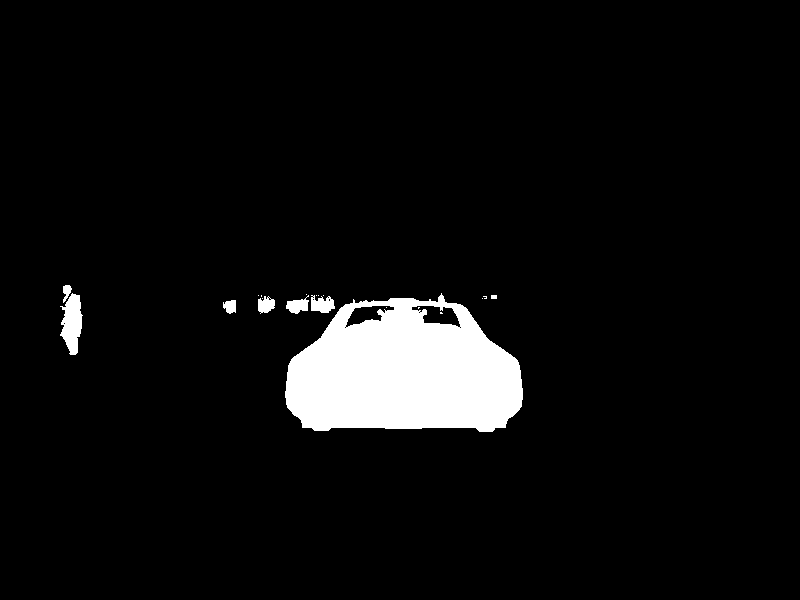}}\end{overpic}}
\hspace{\fill}
\subfloat{\includegraphics[width=.161\linewidth]{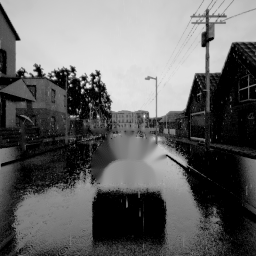}}
\hspace{\fill}
\subfloat{\includegraphics[width=.161\linewidth]{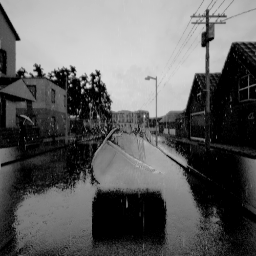}}
\hspace{\fill}
\subfloat{\includegraphics[width=.161\linewidth]{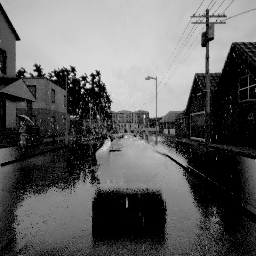}}
\hspace{\fill}
\subfloat{\includegraphics[width=.161\linewidth]{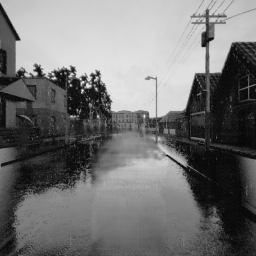}}
\hspace{\fill}
\subfloat{\includegraphics[width=.161\linewidth]{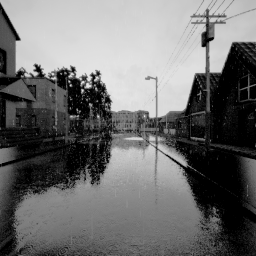}}
\\
\vspace*{-0.6\baselineskip}
\setcounter{subfigure}{0}
\subfloat[\label{fig:input} Input]{\begin{overpic}[width=.161\linewidth]{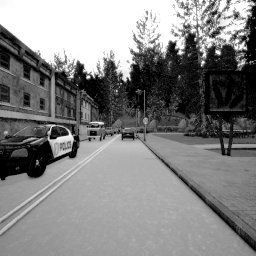}\put(1,56.0){\includegraphics[scale=0.04]{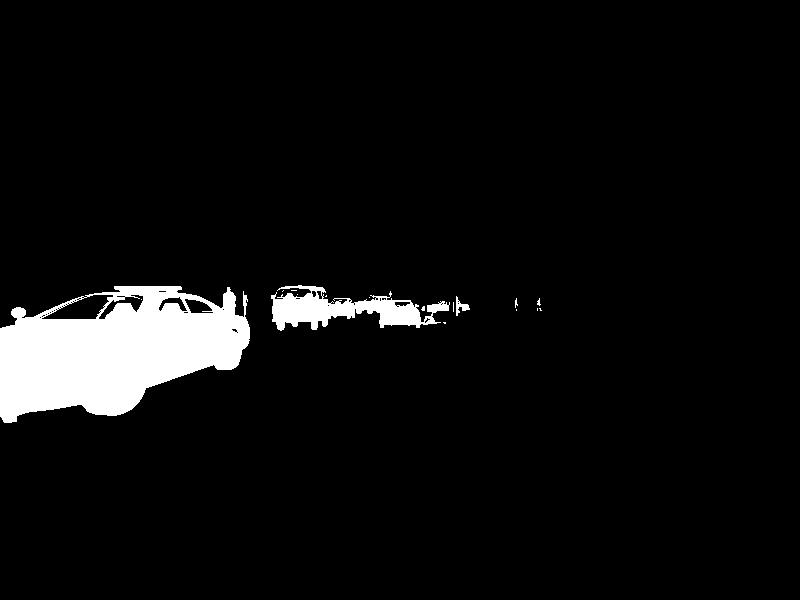}}\end{overpic}}
\hspace{\fill}
\subfloat[\label{fig:output_Geo} Geo \cite{telea2004image}]{\includegraphics[width=.161\linewidth]{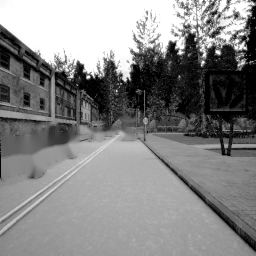}}
\hspace{\fill}
\subfloat[\label{fig:output_Lea1} Lea1 \cite{yu2018generative}]{\includegraphics[width=.161\linewidth]{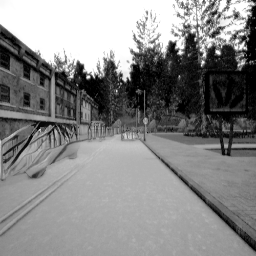}}
\hspace{\fill}
\subfloat[\label{fig:output_Lea2} Lea2 \cite{iizuka2017globally}]{\includegraphics[width=.161\linewidth]{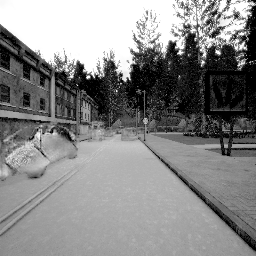}}
\hspace{\fill}
\subfloat[\label{fig:output_ours} Ours]{\includegraphics[width=.161\linewidth]{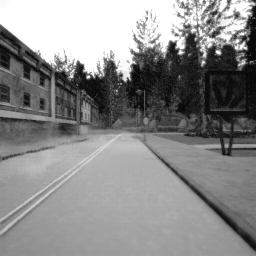}}
\hspace{\fill}
\subfloat[\label{fig:gt} Ground-truth]{\includegraphics[width=.161\linewidth]{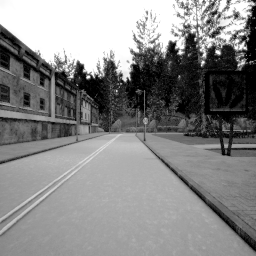}}
\caption{\label{fig:comparisonCARLA} Qualitative comparison of our method \protect\subref{fig:output_ours} against other inpainting techniques \protect\subref{fig:output_Geo}, \protect\subref{fig:output_Lea1}, \protect\subref{fig:output_Lea2} on our synthetic dataset.}
\vspace{-5mm}
\end{figure*}

\subsection{Inpainting Comparisons}
We compare qualitatively and quantitatively our ``inpainting'' method with three other approaches:
\begin{itemize}[leftmargin=5.0mm]
\setlength\itemsep{0.0cm}
\item{\textbf{Geo:}} a state-of-the-art non-learning based approach~\cite{telea2004image}.
\item{\textbf{Lea1}, \textbf{Lea2}:} two deep learning based methods \cite{yu2018generative, iizuka2017globally}.
\end{itemize}

\begin{figure*} [t]
\vspace{-5mm}
\centering
\subfloat{\begin{overpic}[width=.161\linewidth]{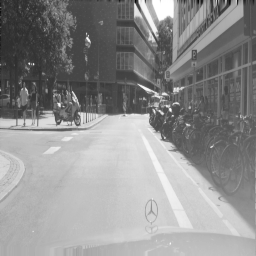}\put(1,56.0){\includegraphics[scale=0.025]{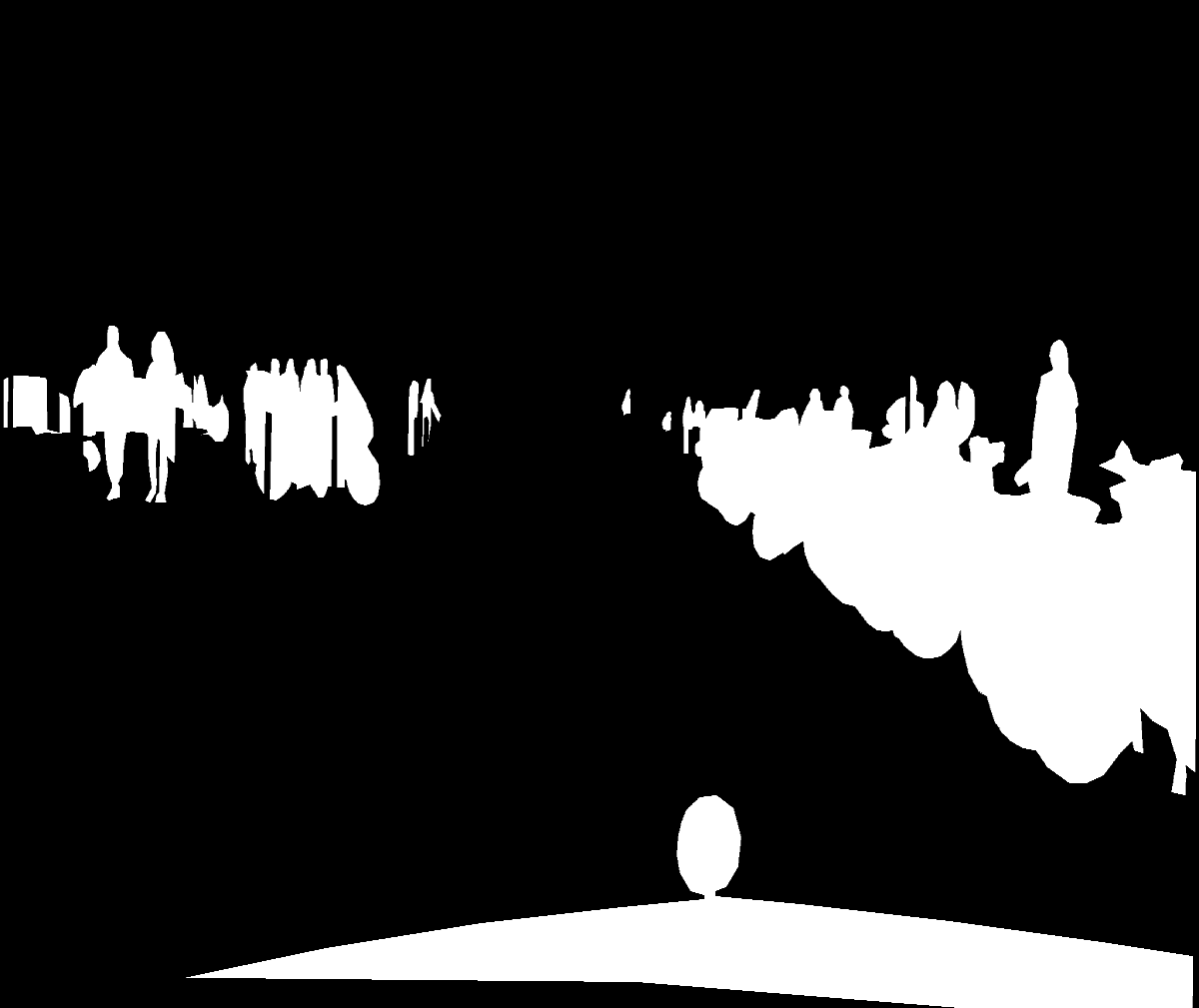}}\end{overpic}}
\hspace{\fill}
\subfloat{\includegraphics[width=.161\linewidth]{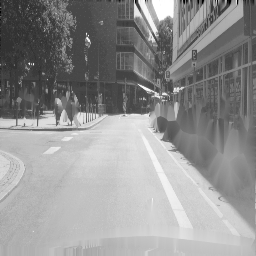}}
\hspace{\fill}
\subfloat{\includegraphics[width=.161\linewidth]{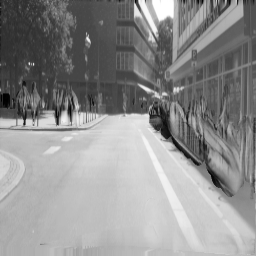}}
\hspace{\fill}
\subfloat{\includegraphics[width=.161\linewidth]{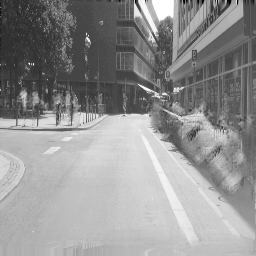}}
\hspace{\fill}
\subfloat{\includegraphics[width=.161\linewidth]{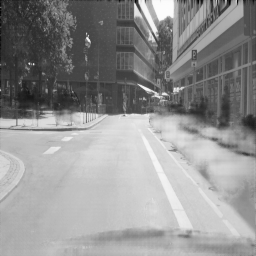}}
\hspace{\fill}
\subfloat{\includegraphics[width=.161\linewidth]{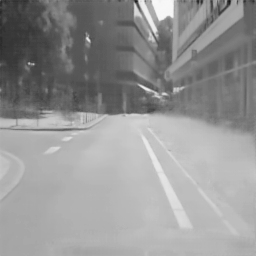}}
\\
\vspace*{-0.6\baselineskip}
\setcounter{subfigure}{0}
\subfloat[\label{fig:nsynth_input}Input]{\begin{overpic}[width=.161\linewidth]{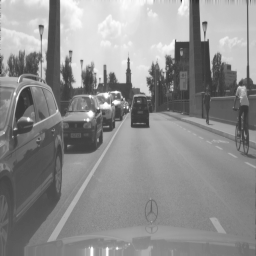}\put(1,56.0){\includegraphics[scale=0.025]{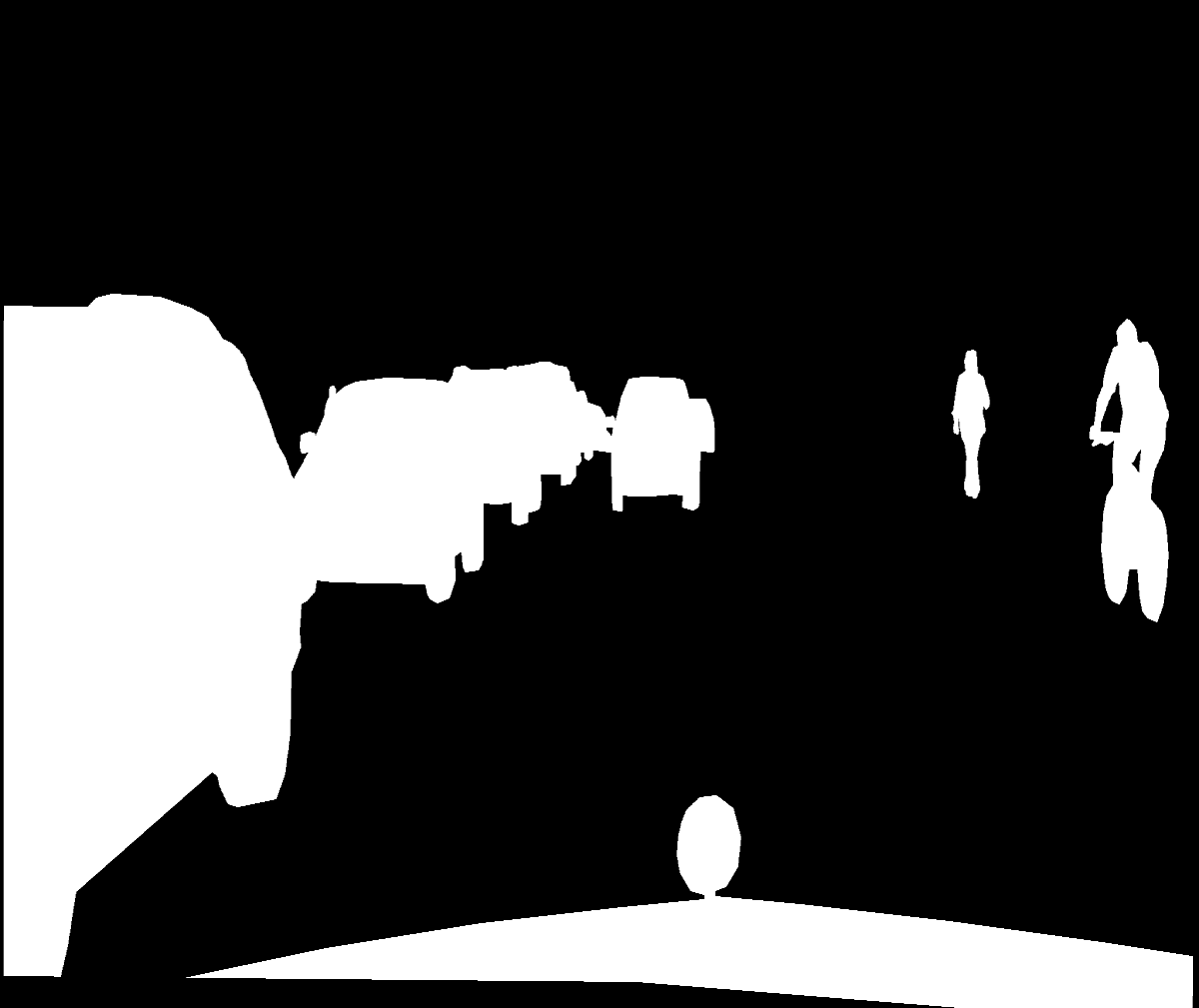}}\end{overpic}}
\hspace{\fill}
\subfloat[\label{fig:nsynth_output_Geo}Geo~\cite{telea2004image}]{\includegraphics[width=.161\linewidth]{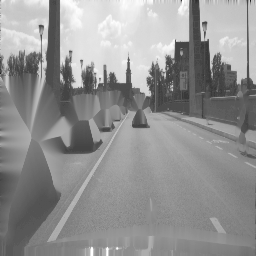}}
\hspace{\fill}
\subfloat[\label{fig:nsynth_output_Lea1}Lea1~\cite{yu2018generative}]{\includegraphics[width=.161\linewidth]{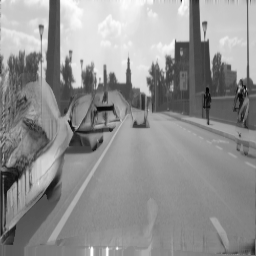}}
\hspace{\fill}
\subfloat[\label{fig:nsynth_output_Lea2}Lea2~\cite{iizuka2017globally}]{\includegraphics[width=.161\linewidth]{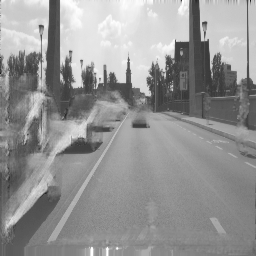}}
\hspace{\fill}
\subfloat[\label{fig:SynthToReal}S$\rightarrow$Real]{\includegraphics[width=.161\linewidth]{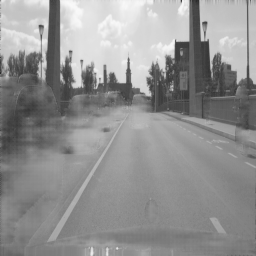}}
\hspace{\fill}
\subfloat[\label{fig:Synth+RealToReal}S+R$\rightarrow$Real]{\includegraphics[width=.161\linewidth]{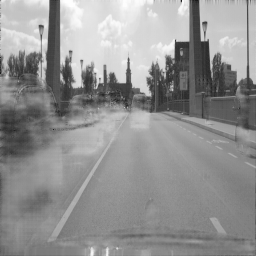}}
\caption{\label{fig:comparisonCITYSCAPES} Qualitative comparison of our method \protect\subref{fig:SynthToReal}, \protect\subref{fig:Synth+RealToReal} against other image inpainting approaches \protect\subref{fig:nsynth_output_Geo}, \protect\subref{fig:nsynth_output_Lea1}, \protect\subref{fig:nsynth_output_Lea2} on the Cityscapes validation dataset~\cite{cordts2016cityscapes}. 
\protect\subref{fig:SynthToReal} shows our results when the training images are all synthetic (S). 
Albeit, \protect\subref{fig:Synth+RealToReal} shows our results when real images from the Cityscapes dataset have been incorporated into our training set together with the synthetic CARLA images with a ratio of $1/10$ (S+R).}
\vspace{-3mm}
\end{figure*}

Since both \textbf{Lea1} and \textbf{Lea2} are methods conceived for general inpainting purposes, we directly use their released models \cite{yu2018generative, iizuka2017globally} trained on the Places2 dataset \cite{zhou2017places}.
We provide them with the same mask than to our method to generate the holes in the images. 
We evaluate qualitatively on the 3000 images from our synthetic test dataset, and on the 500 validation images from the Cityscapes dataset \cite{cordts2016cityscapes}. 
We can see in Figs.~\ref{fig:comparisonCARLA} and~\ref{fig:comparisonCITYSCAPES} the qualitative comparisons on both datasets respectively\footnote{Results generated with both inpainting methods \textbf{Lea1} and \textbf{Lea2} have been generated with the color images at a $256 \times 256$ resolution and then converted to gray scale for visual comparison with our network's output.}. 
%
%
Visually, we observe that our method obtains a more realistic output. 
Also, it is the only one capable of removing the shadows generated by the dynamic objects even though they are not included in the dynamic/static mask (Fig.~\ref{fig:comparisonCARLA} row 2).
The utilized masks are included in the images in Fig. \ref{fig:input}.
Table \ref{tab:comparison} shows the quantitative comparison of our method against \textbf{Geo}, \textbf{Lea1} and \textbf{Lea2} on our CARLA dataset. 
For a fair comparison we only report the $L1$ error within the mask $L1_{in}$.
Reporting the non-hole regions error would be unfair since the other methods are not conceived to notoriously change them.

Regarding the Cityscapes dataset evaluation, quantitatively measuring the performance of the different methods is not possible since ground-truth does not exist. 
In view of these results, we claim that our approach outperforms both qualitatively and quantitatively the other methods in such task.

\begin{table} [t]
\vspace{+3mm}
\begin{tabularx}{\linewidth}{@{}l*{4}{C}}
\toprule
Experiment & Geo \cite{telea2004image} & Lea1 \cite{yu2018generative} & Lea2 \cite{iizuka2017globally} & Ours \\
\midrule
$L1_{in} (\%)$ & 6.66 & 10.45 & 10.49 & \textbf{6.00} \\
\bottomrule
\end{tabularx}
\caption{\label{tab:comparison} Quantitative results of our method against other inpainting approaches in our CARLA dataset.}
\vspace{-5mm}
\end{table}

\subsection{Transfer to Real Data}
Models trained on synthetic data can be useful for real world vision tasks~\cite{gaidon2016virtual, peris2012towards, skinner2016high, tobin2017domain}. 
Accordingly, we provide a preliminary study of synthetic-to-real transfer learning using data from the Cityscapes dataset \cite{cordts2016cityscapes}, which offers a variety of real-world environments similar to the synthetic ones.

When testing our method on real data, we see qualitatively that results are not as good as with the CARLA images (Fig.~\ref{fig:SynthToReal}). 
This happens because such data has different statistics than the real one, and therefore cannot be easily used. 
The combination of real and synthetic data is possible during training despite the lack of ground-truth static real images. 
In the case of the real images, the network only learns the texture and the style of the static real world by encoding its information and decoding back the original image non-hole regions. 
The synthetic data is substantially more plentiful and has information about the inpainting process. 
The rendering, however, is far from realistic. 
Thus, the chosen representation attempts to bridge the reality gap encountered when using simulated data, and to remove the need for domain adaptation. 
Fig.~\ref{fig:Synth+RealToReal} shows how adding real images in the training process leads the testing in real data to give slightly better results\footnote{We perform extensive data augmentation --Gaussian blur, Gaussian noise, brightness, contrast and saturation-- to avoid overfitting to the synthetic data.}. 
%
Still, the results are not as accurate/realistic as the ones obtained with CARLA images.

Differently, using a CycleGAN type approach~\cite{CycleGAN2017} would allow us to work with real-world imagery and hence delete this domain adaptation requirement. 
This approach would learn our desired mapping in the absence of paired images.

\subsection{Visual Localization Experiments}
We believe that the images generated by our framework have a potential use for visual localization tasks. 
Even though utilizing only the static parts of images would also bring benefits to localization systems such as ORB-SLAM~\cite{mur2015orb}, DSO~\cite{engel2018direct}, SVO~\cite{forster2014svo}, \textit{etc.}, they would require modifications, as for example in DynaSLAM~\cite{bescos2018dynaslam} among others~\cite{wang2014motion, sun2018motion}.
Using inpainted images rather than just masked images allows us to use whichever localization system with no modification. 
This is a remarkable strength of our framework.
As a proof of concept, we conduct three additional experiments.

First, we generated a CARLA dataset consisting of 20 different locations with 6 images taken per location. 
These 6 images show a different dynamic objects setup (Fig.~\ref{fig:overview}).
Then the global descriptors (from an off-the-shelf CNN~\cite{olid2018single}) computed from the different versions of the same location were compared. 
The euclidean distance between the descriptors of the scenes with dynamic objects was always greater than that of the images after dynamic removal and inpainting. 
A 32\% average reduction in the distance was observed. 

In the second experiment, we generated 6 CARLA images at 6 different locations with a very similar vehicle setup. 
With the same global descriptor used in the previous experiment, we compared the distances between all possible image pair combinations. 
Then, we obtained the inpainted images with our framework, and computed the same distances. 
We repeated this experiment 4 times varying the vehicle setup used and saw that the mean distance of the inpainted sets was higher than that of the original images by 65\%.

\begin{figure}
    \vspace{-1mm}
    \centering
    \subfloat[\label{fig:SVS_RIn}Ref]{\includegraphics[width=.24\linewidth]{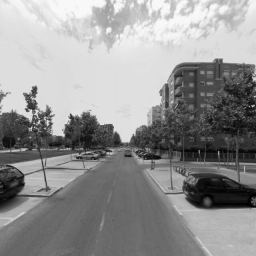}}
    \hspace*{0.001\linewidth}
    \subfloat[\label{fig:SVS_QIn}Query]{\includegraphics[width=.24\linewidth]{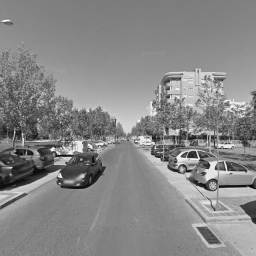}}
    \hspace*{0.01\linewidth}
    \subfloat[\label{fig:SVS_ROut}Empty~Ref]{\includegraphics[width=.24\linewidth]{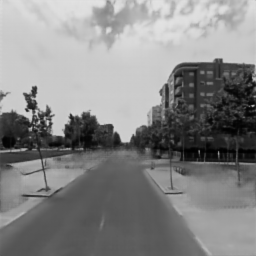}}
    \hspace*{0.001\linewidth}
    \subfloat[\label{fig:SVS_QOut}Empty~Query]{\includegraphics[width=.24\linewidth]{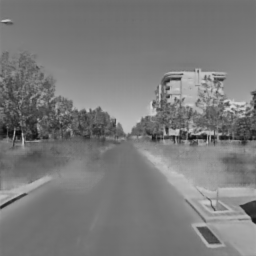}}
    \caption{\label{fig:SVS}\protect\subref{fig:SVS_RIn} and \protect\subref{fig:SVS_QIn} show the same location at different times with different viewpoints, weather conditions and dynamic objects setups~\cite{werburg2018street}. 
    The place recognition work by Olid \textit{et~al.}~\cite{olid2018single} fails to match them as the same place. 
    However, it succeeds in matching them when our framework is previously employed --\protect\subref{fig:SVS_ROut} and \protect\subref{fig:SVS_QOut}--.}
    \vspace{-4mm}
\end{figure}

The third experiment was conducted with real world images from the SVS dataset~\cite{werburg2018street}. 
We performed place recognition~\cite{olid2018single} with both the original images, and the ones processed by our framework. 
In the first case, this task was successful in 58\% of the cases, whereas with our images the success rate was of 67\%. 
Fig.~\ref{fig:SVS} shows a case in which our framework makes place recognition successful.
Even though the inpainting algorithm is not perfect and might introduce false appearance, the two images global descriptors are closer with the fake static images than with the dynamic ones.

In view of these results, our framework brings closer images from the same place with different dynamic objects while pulling apart images from different places but with similar dynamic objects. 
We are confident that localization and mapping systems could benefit from these advances. 
Also, we expect similar methods to show comparable improvements by incorporating our proposal. 
An extended version of this work would include such inquiries. 
Furthermore, a strong benefit of our approach is that such methods would require no modification to work with our processed images.

\subsection{Timing Analysis}
Reporting our framework efficiency is crucial to judge its suitability for robotic tasks.
The end-to-end pipeline runs at 50~fps on a nVidia GeForce GTX 1070 8GB with images of a 256$\times$256 resolution. 
Out of the 20~ms it takes to process one frame, 18~ms are invested into obtaining its \ac{ss}, and 2~ms are used for the inpainting task. 
Other than to deal with dynamic objects, the \ac{ss} may be needed for many other tasks involved in automatic navigation. 
In such cases, our framework would only add 2 extra ms per frame. 
Based on our analysis, we consider that the inpainting task is not the bottleneck, even though higher resolution images may be needed.

\section{CONCLUSION}
\label{sec:conclusion}
We have presented an end-to-end deep learning framework that takes as input an RGB image from a city environment containing dynamic objects such as cars, and converts it into a gray realistic image with only static content. 
For this objective, we develop mGANs, an adaptation of generative adversarial networks for inpainting problems.
The provided comparison against other state-of-the-art inpainting methods shows that our approach performs better. 
Also, our approach has a feature that makes it different from other inpainting methods: areas of the non-hole image can be changed for the objective of a more realistic output.
Further experiments show that visual localization and mapping systems can benefit from our advances without any further modification.

Future work might include, among others, converting the resulting static images from gray scale to \mbox{color~\cite{li2018closed, IizukaSIGGRAPH2016}}. 
Also, exploiting higher-resolution models would be convenient for robotic labours, as well as carrying out more research efforts on its transferability to the real world domain.

\addtolength{\textheight}{-2.5cm}   


\bibliographystyle{ieeetr}
\bibliography{ref}  

\end{document}